\documentclass[onecolumn,journal]{IEEEtran}

\usepackage{color}

\usepackage{algorithmic}
\usepackage{hyperref} 
\usepackage{multirow}
\usepackage{booktabs}

\begin{document}

\title{A Novel  Genetic Algorithm using Helper Objectives for the 0-1 Knapsack Problem }
 \author{Jun He,  Feidun He and  Hongbin Dong 
\thanks{This work was supported by the EPSRC under Grant No. EP/I009809/1.}
\thanks{Jun He   is with Department of Computer Science, Aberystwyth University, Aberystwyth, SY23 3DB, UK (email:  jun.he@aber.ac.uk).}%
\thanks{ Feidun He is with School of Information Science and Technology, Southwest Jiaotong University,
Chengdu, China}
\thanks{Hongbin Dong is with College of Computer Science and Technology, Harbin Engineering University,
Harbin,   China  }

}
\maketitle

\begin{abstract}
The 0-1 knapsack problem is a well-known combinatorial optimisation problem. Approximation algorithms have been designed for solving it and they  return provably good solutions   within  polynomial time.  On the other hand, genetic algorithms are well suited for solving the knapsack problem and they   find reasonably good solutions quickly. A naturally arising question is whether genetic algorithms are able to find  solutions as good as approximation algorithms do. This paper presents   a novel multi-objective optimisation genetic algorithm for solving the 0-1 knapsack problem.       Experiment results show that  the new algorithm outperforms its rivals, the greedy algorithm, mixed strategy genetic algorithm,  and  greedy algorithm + mixed strategy genetic algorithm.

 \end{abstract}

\section{Introduction}
The 0-1 knapsack problem is one of the most important and also  most intensively studied  combinatorial optimisation problems \cite{martello1990knapsack}. 
Several approximation algorithms have proposed for solving the 0-1 knapsack problem \cite{martello1990knapsack}.  These algorithms always can return  provably good solutions, whose values are within a factor  of the value of the optimal solution.

In   last two decades, evolutionary algorithm, especially genetic algorithms (GAs), have been well adopted for tackling the knapsack problem~\cite{michalewicz1994genetic,khuri1994zero}.
The problem has received a particular interest from the evolutionary computation community for the following  reason. The binary vector representation  is a natural encoding of of the candidate solutions to the 0-1 knapsack problem. Thereby, it provides an ideal setting for the  applications of GAs~\cite[Chapter 4]{michalewicz1996genetic}.  

Empirical results often assert that GAs  produce reasonably good solutions to the knapsack problems~\cite{zitzler1999multiobjective,jaszkiewicz2002performance,eugenia2003solving}. A naturally arising question is to compare the solution quality (reasonably good versus provably good)  between GAs  and approximation algorithms.     There are two approaches to answer the question. One approach is to make a theoretical analysis. A GA is proven that it can produce a solution within a polynomial runtime, the value of which is within a factor of  the value of an optimal solution. This is a standard approach used in the study of approximation algorithms. 
Another approach is to conduct an empirical study. A GA  is compared with an  approximation algorithm via computer experiments. If the GA can produce solutions better or not worse than an approximative algorithm does in all instances within polynomial time,   the GA  is able to reach the same solution quality as the approximation algorithm does.  

The current paper is   an empirical  study of an GA which uses the \emph{multi-objectivization} technique~\cite{knowles2001reducing}.  In multi-objectivization, single-objective optimisation problems are transferred  into multi-objective optimisation problems by decomposing the original objective into several components~\cite{knowles2001reducing} or by adding  helper objectives~\cite{jensen2005helper}. Multi-objectivization may bring both positive and  negative  effects~\cite{handl2008multiobjectivization,brockhoff2009effects,lochtefeld2011helper}.   This approach has been used for solving several combinatorial optimisation problems,  for example, the  knapsack problem \cite{kumar2006analysis}, vertex cover problem \cite{friedrich2010approximating} and minimum label spanning tree problem \cite{lai2014performance}.  
This paper focusses on the 0-1 knapsack problem.  
A novel GA using three helper objectives is designed for solving the 0-1 knapsack problem. Then the solution  quality of  the GA is compared with a well-known approximation algorithm via computer experiments.

The remainder of the paper is organized as follows.  The 0-1 knapsack problem, a greedy algorithm and a GA for it are introduced in Section \ref{secProblem}. In  Section~\ref{secGAs} we present a novel GA using helper objectives. Section~\ref{secExperiments} is devoted to an empirical comparison among several algorithms.  Section~\ref{secConclusions} concludes the article.

 \section{Knapsack Problem, Greedy Algorithm and Genetic Algorithm} \label{secProblem}
In an instance of the 0-1 knapsack problem, given a set of $n$ items with weights
$w_i$ and profits $p_i$, and a knapsack with capacity $C$, the task is to maximise the sum of profits of items packed in the knapsack without exceeding the capacity. More formally the target is to find a binary vector $\vec{x}=(x_1 \cdots x_n)$ so as to
\begin{equation}
\label{equ:single-objective}
\max_{\vec{x}}   f(\vec{x})= \sum^n_{i=1} p_i x_i,   
\quad  \mbox{subject to } \sum^n_{i=1} w_i x_i \le C,
\end{equation}
where  $
x_i =   1$ if the item $i$ is
selected in the knapsack and $x_i=
0$ if the item $i$ is not selected in the knapsack.  
A feasible solution is  an $\vec{x}$ which satisfies the constraint. An infeasible one is an $\vec{x}$ that violates the constraint.  

Several approximation algorithms have been proposed for solving the 0-1 knapsack problem (see \cite[Chapter 2]{martello1990knapsack} for more details). Among these, the simplest one is the greedy algorithm described below. The algorithm aims at putting the most profitable items as many as possible into  the knapsack or the items with the highest profit-to-weight ratio as many as possible, within the knapsack capacity.
 \begin{algorithmic} [1]
\STATE \textbf{input} an instance of the 0-1 knapsack problem;
 \STATE resort all the items via the ratio of their profits to their corresponding weights so that $\frac{p_1}{w_1} \ge \cdots \ge \frac{p_n}{w_n}$;

\STATE greedily add the items in the above order to the knapsack as long as adding an item to the knapsack does not exceeding the capacity of the knapsack. Denote the solution by $\vec{y}$;

\STATE resort all the items according to their profits so that ${p_1}\ge {p_2} \ge \cdots \ge {p_n}$;

\STATE greedily add the items in the above order as long as adding an item to the knapsack does not exceeding the capacity of the knapsack. Denote the solution by $\vec{z}$;
\STATE \textbf{output} the best of $\vec{y}$ and $\vec{z}$.
\end{algorithmic}
 
This algorithm is a $1/2$-approximation algorithm for the 0-1 knapsack problem \cite[Section 2.4]{martello1990knapsack}, which means it always can return a solution no  worse than 1/2 of the value of the optimal solution. 

The  greedy algorithm stops after finding an approximation solution, and it has no ability to  seek the global optimal solution.  Therefore  GAs are often applied for solving the 0-1 knapsack problem.

In order to handle the constraint in the knapsack problem, we use  repair methods since they are claimed to be the most efficient  for the  knapsack problem~\cite{michalewicz1996genetic,he2007comparison}. 
A repair method  is explained as follows.
\begin{algorithmic}[1]
\STATE \textbf{input} an infeasible solution $\vec{x}$;
\WHILE{$\vec{x}$ is infeasible}
\STATE $i=:$ \textbf{choose} an item from the knapsack;
\STATE set $ x_i=0$;
\IF{$\sum^n_{i=1} x_i w_i \le C$}
\STATE $\vec{x}$ is feasible;
\ENDIF
\ENDWHILE
   
\STATE \textbf{output} a feasible solution $\vec{x}$.
\end{algorithmic}

There are different  methods available for \emph{choosing}  an item in the \emph{repair} procedure,  described as follows.
\begin{enumerate}
\item
\emph{Profit-greedy repair:}
sort all items  according to the decreasing order of their corresponding profits. Then choose the item with the smallest profit and remove it from the knapsack.
\item
\emph{Ratio-greedy repair:}
sort all items  according to the decreasing order of the corresponding ratios. Then choose the item with the smallest ratio and remove it from the knapsack.

\item
\emph{Random repair:} choose an item  from the knapsack at random and remove it from the knapsack.
\end{enumerate}

Thanks to the repair method, all of the infeasible solutions are repaired into the feasible ones. The following pseudo-code is a mixed strategy  GA (MSGA) which chooses one of three repair methods in a probabilistic way and then applies the repair method to generate a feasible solution.

  \begin{algorithmic} [1]
\STATE \textbf{input} an instance of the 0-1 knapsack problem;
\STATE initialize population $ \Phi_0$ consisting of $N$ feasible solutions;
 
\FOR{$t=0,1, \cdots, t_{\max} $ }
\STATE generate a random number $r$ in $[0,1]$;
\IF{$r<0.9$} 
\STATE children population $\Phi_{t.c}$ $\leftarrow$   bitwise-mutate  $\Phi_t$;
\ELSE
\STATE  children population $\Phi_{t.c}$ $\leftarrow$   one-point crossover $\Phi_t$;
\ENDIF
\IF{a  child  is an infeasible solution}
\STATE choose one method from the ratio-greedy repair, random repair and value-greedy repair with probability 1/3 and repair the child into   feasible;
\ENDIF
 
\STATE the best individual in the parent and children populations is selected into population $\Phi_{t+1}$; 
\STATE    $N-1$ individuals from the parent and  children populations into population $\Phi_{t+1}$ by roulette wheel  selection; 
\ENDFOR
\STATE \textbf{output} the maximum of the fitness function.
\end{algorithmic} 
The   genetic operators used in the   above GA are explained below. 
\begin{itemize}
\item \emph{Bitwise Mutation:}  Given a binary vector $(x_1  \cdots  x_n)$, flip each bit $x_i$    with probability $ {1}/{n}$.
\item \emph{One-Point Crossover:} Given two binary vectors  
$
 (x_1 \cdots x_n)$ and $(y_1 \cdots y_n),
$
 randomly choose a crossover point $k \in \{1, \cdots, n\}$, swap their bits at point $k$. Then generate two new binary vectors as follows,
$$
 (x_1 \cdots x_k y_{k+1} \cdots y_n), \quad   (y_1 \cdots y_k x_{k+1} \cdots x_n). 
$$
\end{itemize}

Like most of GAs, the MSGA may find reasonably good solutions but has no guarantee about the solution quality.   Thus it is necessary to  design evolutionary approximation algorithms with provably good solution quality. The most straightforward approach is that we first apply the greedy algorithm for generating approximation solutions and then take these solutions as the  starting point of the MSGA.  We call this approach \emph{greedy algorithm + MSGA}.  Since the MSGA starts at local optima, it becomes hard for the MSGA to leave the absorbing basin of this local optimum for seeking the global optimum. This is the main drawback of the approach.  

\section{Genetic Algorithm using Helper Objectives for the 0-1 Knapsack Problems} 
\label{secGAs}

In this  section, we propose a novel multi-objective optimisation GA (MOGA)  which can beat the   combination of greedy algorithm + MOGA mentioned in the previous section. The algorithm is based on  the  {multi-objectivization} technique.
The original single objective optimization problem (\ref{equ:single-objective}) is recast into a multi-objective optimization problem using helper objectives.   The design of helper objectives  depends on problem-specific knowledge.
The first helper objective comes from an observation on the following instance. 
\begin{center}
\begin{tabular}{c|c|c|c|c|c}
\toprule
Item  & 1  & 2 & 3 & 4 &5 \\
\midrule
Profit   &  10 & 10  &  10 & 12  & 12\\
Weight   & 10  & 10 & 10   & 10  &10 \\
\midrule
Capacity &\multicolumn{5}{c}{20}\\
\bottomrule
\end{tabular}
\end{center}

The global optimum in this instance  is  
 $00011$.   In the optimal solution, the  average profit of packed items is the largest. Thus the first helper objective is to maximize the average profit of items in a knapsack.    The objective function is
\begin{equation}
   h_1(\vec{x}) = \frac{1}{\parallel \vec{x} \parallel_1}\sum^n_{i=1}   x_i p_i.
\end{equation}
where $\parallel \vec{x} \parallel_1 =\sum^n_{i=1} x_i$.

The second objective is inspired from an observation on another instance.
\begin{center}
\begin{tabular}{c|c|c|c|c|c}
\toprule
Item  & 1  & 2 & 3 & 4 &5 \\
\midrule
Profit   &  15 & 15  &  20 & 20  & 20\\
Weight   & 10  & 10 & 20   & 20  &20 \\
\midrule
Capacity &\multicolumn{5}{c}{20}\\
\bottomrule
\end{tabular}
\end{center}

The global optimum in this instance  is  
 $11000$.   In the optimal solution, the  average  profit-to-weight ratio of packed items is the largest. However, the average profit of these items is not the largest. Then  the second helper objective is to maximize the average profit-to-weight ratio of items in a knapsack.   The  objective function is
\begin{equation}
   h_2(\vec{x}) = \frac{1}{\parallel \vec{x} \parallel_1}\sum^n_{i=1}   x_i \frac{p_i}{w_i}.
\end{equation} 

Finally let's look at the following instance.
\begin{center}
\begin{tabular}{c|c|c|c|c|c}
\toprule
Item  & 1  & 2 & 3 & 4 &5 \\
\midrule
Profit   &  40 & 40  &  40 & 40 & 150\\
Weight   & 30 & 30&30   & 30  &100 \\
\midrule
Capacity &\multicolumn{5}{c}{120}\\
\bottomrule
\end{tabular}
\end{center}
 
It is not difficult  to verify that the global optimum in this instance  is  
 $11110$.   In the optimal solution,  neither the average profit of packed items nor average profit-to-weight ratio  is  the largest, but the  number of packed items is the largest. Thus the  third helper objective is to maximize the number of items in a knapsack.   The objective function  is
\begin{equation}
   h_3(\vec{x}) = \parallel \vec{x} \parallel_1 .
\end{equation}

 We then come to the following multi-objective optimization problem:
\begin{equation}
\label{equ:multi-objective} 
 \max_{\vec{x}} \{ f(\vec{x}),  h_1(\vec{x}) , h_2(\vec{x}),   h_3(\vec{x}) \}, 
  \qquad  \mbox{subject to } \sum^n_{i=1} w_i x_i \le C.
\end{equation}

Besides the above three helper objectives, it is possible to add more helper objectives, for example, to minimise the average weight of packed items.

The multi-objective optimisation problem (\ref{equ:multi-objective})  is solved by  a MOGA using  bitwise mutation, one-point crossover and multi-criteria selection, plus a mixed strategy of three repair methods.

 \begin{algorithmic} [1]
\STATE \textbf{input} an instance of the 0-1 knapsack problem;
\STATE initialize $ \Phi_0$ consisting of $N$ feasible solutions;
 \FOR{$t=0,1, \cdots, t_{\max}$ }
\STATE generate  a random number $r$ in $[0,1]$;
\IF{$r<0.9$} 
\STATE children population $\Phi_{t.c}$ $\leftarrow$   \emph{bitwise mutate}  $\Phi_t$;
\ELSE
\STATE  children population $\Phi_{t.c}$ $\leftarrow$   \emph{one-point crossover }$\Phi_t$;
\ENDIF
\IF{any  child  is an infeasible solution}
\STATE choose one repair method from the ratio-greedy repair, random repair and value-greedy repair with probability 1/3;
\STATE repair the child into a feasible solution;
\ENDIF
 
 \STATE population $\Phi_{t+1}$ $\leftarrow$  \emph{multi-criterion select} $N$ individuals from $\Phi_t$ and $\Phi_{t.c}$;
\ENDFOR
\STATE \textbf{output} the maximum of $f(\vec{x})$ in the final population.
\end{algorithmic} 

 The \emph{multi-criteria selection} operator, adopted in the above MOGA, is novel and inspired from multi-objective optimisation. Since the target is to maximise several objectives simultaneously, we  select individuals which have higher function values with respect to each objective function.  The pseudo-code of multi-criteria selection is described as follows.

\begin{algorithmic} [1]
\STATE \textbf{input} the parent population $\Phi_t$  and   child population $\Phi_{t.c}$;
 
\STATE merge the parent and children populations into a temporary population   which consists of $2N$ individuals;

\STATE sort  all individuals in  the temporary populations  in the descending order of $f(\vec{x})$, denote them by $\vec{x}^{(1)}_1, \cdots, \vec{x}^{(1)}_{2N}$; 
 
\STATE select all individuals from left to right (denote them by $\vec{x}^{(1)}_{k_1}, \cdots, \vec{x}^{(1)}_{k_m}$) which satisfy $h_1(\vec{x}^{(1)}_{k_i})< h_1(\vec{x}^{(1)}_{k_{i+1}})$ or $h_2(\vec{x}^{(1)}_{k_i})< h_2(\vec{x}^{(1)}_{k_{i+1}})$ for any $k_i$. 
\IF{the   number of selected individuals  is greater than $m\frac{N}{3}$}
\STATE truncate them to $\frac{N}{3}$ individuals;
\ENDIF

\STATE add these selected individuals into the next population $\Phi_{t+1}$;

\STATE resort  all individuals in  the temporary population  in the descending order of  $h_1(\vec{x})$,  still denote them by $\vec{x}_1, \cdots, \vec{x}_{2N}$;    

\STATE select all individuals from left to right (still denote them by $\vec{x}_{k_1}, \cdots, \vec{x}_{k_m}$) which satisfy $h_3(\vec{x}_{k_i})< h_3(\vec{x}_{k_{i+1}})$  for any $k_i$. 
\IF{ the   number of selected individuals is greater than $\frac{N}{3}$}
\STATE truncate them to $\frac{N}{3}$ individuals;
\ENDIF
\STATE add these selected individuals into the next population $\Phi_{t+1}$; 

\STATE resort  all individuals in  the temporary populations  in the descending order of $h_2(\vec{x})$,  still denote them by $\vec{x}_1, \cdots, \vec{x}_{2N}$; 
 
\STATE select all individuals from   left to right (still denote them by $\vec{x}_{k_1}, \cdots, \vec{x}_{k_m}$) which satisfy $h_3(\vec{x}_{k_i})< h_3(\vec{x}_{k_{i+1}})$ for any $k_i$. 
\IF{ the   number of selected individuals is greater than  $\frac{N}{3}$}
\STATE truncate them to $\frac{N}{3}$ individuals;
\ENDIF
\STATE add these selected individuals into the next population $\Phi_{t+1}$; 

\WHILE{the population size of $\Phi_{t+1}$ is less than  $N$} 
\STATE randomly choose an individual from the parent population  and add it into $\Phi_{t+1}$;
\ENDWHILE

\STATE \textbf{output} a new population $\Phi_{t+1}$.
\end{algorithmic}

In the above algorithm, Steps 3-4 are for selecting the individuals  with higher values of $f(\vec{x})$. In order to preserve  diversity, we  choose these individuals which have different values of $h_1(\vec{x})$ or $h_2(\vec{x})$. Similarly
Steps 9-10 are for selecting the individuals with a higher value of $h_1(\vec{x})$. We  choose the individuals which have different values of $h_3(\vec{x})$ for maintaining diversity.
Steps 15-16 are for selecting individuals with a higher value of $h_{2}(\vec{x})$. Again we  choose these individuals which have different values of $h_3(\vec{x})$ for preserving diversity. 
We don't explicitly select individuals based on  $h_3(\vec{x})$. Instead we implicitly do it  during  Steps 9-10, and Steps 15-16.

Steps  5-7, Steps 11-13,  Steps 17-19, plus Steps 21-23 are used to maintain an invariant population size $N$.

The benefit of using multi-criterion selection is its ability of making search along different directions $f(\vec{x}), h_1(\vec{x})$, $h_2(\vec{x})$ and implicitly $h_3(\vec{x})$. Hence the MOGA may  not get trapped into the absorbing area of a local optimum.

 \section{Experiments}
\label{secExperiments}
In this section, we implement computer experiments.   According to~\cite{martello1990knapsack,michalewicz1996genetic}, the instances of the 0-1 knapsack problem are often classified  into two categorises. 
\begin{enumerate}
\item {\it Restrictive capacity knapsack:}  the knapsack capacity  is so
small that only a few items can be packed in the knapsack. 
An instance with restrictive capacity knapsack is generated in the following way.  Choose a parameter $B$ which is an upper bound on the weight of each item. 
In the experiments,  set  $B=n$.   For item $i$, its  profit $p_i$ and weight $w_i$ are generated at uniformly   random in $[1, B]$. Set  the capacity of the knapsack   $C=B$.

\item {\it Average capacity knapsack:}  the knapsack  capacity   is so large that it is possible to pack  half of items into the knapsack. 
 An instance with average capacity knapsack is generated as follows. Choose a parameter $B$ which is the upper bound on the weight of each item.
In the experiments,  set  $B=n$.  For item $i$, its profit $p_i$ and weight $w_i$ are generated at uniformly   random in $[1, B]$. Since the average weight of each item is $0.5 B$, thus the average of the total weight of items is $0.5nB$. So we set the capacity to be the half of the total weight, that is $C=0.25 n   B$.
\end{enumerate}

  For each type of the 0-1 knapsack problem, 10 instances are generated at random. For each instance, the number of items $n$ is $100$.
The population size is $3n$. The number of maximum generations is   $30n$ for the MSGA and $10n$ for the MOGA. All individuals in the initial population are generated at random. If an individual is an infeasible solution, it is repaired to feasible using random repair.

Besides the above randomly generated instances, we also consider two special instances. Special instance I is given Table~\ref{tabInstanceI}.  
\begin{table*}[ht]
\centering
\caption{Special Instance I:  $n=500$ and $\alpha=0.2$}
\label{tabInstanceI} 
\begin{tabular}{c|c|c|c}
\toprule
Item $i$ & $1, \cdots, \lceil\frac{n}{1+\alpha}\rceil$ & $\lceil\frac{n}{1+\alpha}\rceil+1$ & $\lceil\frac{n}{1+\alpha}\rceil+2, \cdots,  n$ \\
\midrule
Profit $p_i$ &  1 &$\frac{ \alpha n}{1+\alpha} $     & $\frac{1}{n} $  \\
Weight $w_i$ &  1  &  $ \frac{n}{1+\alpha}  - \frac{\alpha}{4+4\alpha} $  &$\frac{1}{2n}$     \\
\midrule
Capacity &\multicolumn{3}{c}{$ \frac{n}{1+\alpha} $}\\
\midrule
Initialisation &0 &   1 &  half bits are 1   \\
\bottomrule
\end{tabular}
\end{table*}
 
 Special instance II is given in Table~\ref{tabInstanceII}. 
\begin{table*}[ht]
\centering
\caption{Special Instance II:  $n=200$}
\label{tabInstanceII} 
\begin{tabular}{c|c|c|cc}
\toprule
Item $i$ & $1, \cdots, \frac{n}{4}$ & $\frac{n}{4}+1, \cdots, \frac{n}{2}$ & $\frac{n}{2}+1,\cdots, n$   \\
\midrule
Profit $p_i$ & $0.25n \sqrt{n}+ 2$  & $0.3 n\sqrt{n} $  &  $\sqrt{n}$     \\
Weight $w_i$ &  $ 0.25 n \sqrt{n}+1$   &  $0.5 n \sqrt{n} $   & $\sqrt{n}$      \\
\midrule
Capacity &\multicolumn{3}{c}{$0.5 n \sqrt{n}$ }\\
\midrule
Initialisation &   one bit is 1,   others 0 &0&  half bits are 1 \\
\bottomrule
\end{tabular}
\end{table*}

The population size is $n$ for Instances I and II. The number of maximum generations is $15n$  for the MSGA and  $5n$ for the  MOGA.   The initialisation  of individuals in both MSGA and MOGA refer  to the above tables.

Tables \ref{tabResults} gives experiment results of comparing  the greedy algorithm, MSGA, greedy algorithm + MSGA and MOGA. From the table, we observe that 
\begin{itemize}
\item the solution quality of MSGA is better or not worse than  the greedy algorithm  in 20 random instances. However for  Instance I, the MSGA  only finds a solution whose value is about 20\% of the optimal value. 
\item  the solution quality of greedy algorithm + MSGA is   better or not worse than  the greedy algorithm  in all instances. However for   Instance II, the algorithm gets trapped into a local optimum, and  is worse than the MOGA.
\item   the MOGA is the winner among 4 algorithms and its the solution quality  is   better or not worse than  the greedy algorithm and MSGA in all instances.   
\end{itemize}
 
\begin{table*}[ht]
\centering
\caption{A comparison among 4 algorithms in 20  randomly generated instances and 2 special instances. The first 10 instances belong to the restrictive capacity knapsack problem. The second 10 instances belong to the average capacity knapsack problem. `max': the maximum  value of $f(\vec{x})$ produced during 10 runs. `average': the average value of $f(\vec{x})$ over 10 runs. `stdev': the standard derivation of $f(\vec{x})$ in 10 runs.}
\label{tabResults} 
\begin{tabular}{c|c|ccc|ccc|cccccccc}
\toprule
& Greedy & \multicolumn{3}{c|}{MSGA }& \multicolumn{3}{c|}{Greedy + MSGA }   & \multicolumn{3}{c}{MOGA }\\
\midrule
 Instance  &  & max & average & stdev&  max & average & stdev &  max & average & stdev \\
\midrule
  1 &  674   & \textbf{683} & 683 	& 0  &\textbf{683} &	683 &	0 & \textbf{683}	& 681.2	& 1.55
\\
   2 & \textbf{714}   & \textbf{714} & 714 &	0 & \textbf{714}	& 714	&0 &\textbf{714}	&714&	0
\\
  3 & 561   & \textbf{622 }& 622 &	0 & \textbf{622}	& 622	&0 & \textbf{622}	&622	&0
 \\
  4 & \textbf{631}   & \textbf{631 }& 631	& 0 &\textbf{631}	&631	& 0 & \textbf{631}	& 631&	0
 \\
  5 & 585    & \textbf{621} & 621	& 0 & \textbf{621}	& 620.7 &	0.95 & \textbf{621}	&620.7	& 0.95
\\
  6 &\textbf{ 787 }   & \textbf{787 }&787	& 0 &\textbf{ 787 }&	787 &	0 & \textbf{787 }&	787& 	0

\\
  7 & 736    & \textbf{773 }&773	&0 & \textbf{773}& 	773&	0 & \textbf{773 }&	773 &	0
 \\
    8& 1042     & \textbf{1076 }& 1076	& 0& \textbf{1076} &	1076 &	0 &\textbf{1076}	&1076	&0
 \\
  9 & 982     & \textbf{994} & 994	&0 & \textbf{994}	& 993	& 3.16 & \textbf{994 }&	993	& 3.16
\\
  10 & 906  &\textbf{ 942} & 942 &	0 & \textbf{942} &	942	&0 & \textbf{942	}& 942 &	0
 \\
 \midrule
  11 & 4107  & \textbf{4111}	&4110.9	&0.32
 & \textbf{4111}	&4111	&0 &\textbf{4111}	&4111 &	0
 \\
  12 & 4090   &\textbf{4102}	&4100.6	&1.43
 &\textbf{4102}	& 4097.1	& 4.41 & \textbf{4102}&	4101.2	&1.03
   \\
  13  & 4138 &\textbf{4169}	&4168.6	&1.26
 & \textbf{4169} &	4165.6	& 2.37 & \textbf{4169}	&4169 &	0
  \\
  14 & 3901  &3925	&3923.9	&1.29
 & 3925	& 3922.2 &	2.15 & \textbf{3927}	& 3923.7	& 1.95
 \\
   15 & 3997  &\textbf{4047}	&4047	&0
 & \textbf{4047}	& 4043.2	&3.71 & \textbf{4047}	&4047 &	0
  \\
  16 & 3984   &\textbf{3994}	&3993&	1.05 
 & \textbf{3994}	& 3992.5	& 0.85 & \textbf{3994} &	3993.8	& 0.632
\\
  17 & 3820  & \textbf{3848}	&3848	&0
 &\textbf{3848}	& 3845.1	&1.97 & \textbf{3848}	& 3848 &	0
  \\
  18 & 3914   &\textbf{3920}	&3919.4	&1.90
 & \textbf{3920} &	3915.2	&2.53 &\textbf{3920} &	3920 &	0 \\
  19 &  4456  &\textbf{4471}	& 4470.1	&0.57
 &4470	&4465.4	& 2.59 & \textbf{4471} &	4470.4 &	0.97
   \\
  20 & 4149   &\textbf{4177}	&4175.3	&1.34
 & \textbf{4177}	& 4171.9 &	2.73 & \textbf{4177}	& 4176	& 0.82  \\
\midrule
   I  &   \textbf{  416.2   }& 83.4    &	83.4 	&0 & \textbf{ 416.2     } &	 416.2   &	0
&\textbf{  416.2  }	&  416.2   & 0
 \\
    II  &  1402.1 &  1402.1 & 	1402.1&0 & 1402.1
 &	1402.1 &	0
&\textbf{ 1414.2 
}	& 1414.2  
& 0\\
\bottomrule
\end{tabular}
\end{table*}

\section{Conclusions}
\label{secConclusions}
A novel MOGA using helper objectives is proposed in this paper for solving the 0-1 knapsack problem. First the original 0-1 knapsack problem  is recast into a  multi-objective optimization problem (i.e. to maximize  the sum of profits packed in the knapsack, to maximize the average profit-to-weight ratio of items, to maximize the average profit of items, and  to maximize the number of packed items).
Then a   MOGA (using bitwise mutation,  one-point crossover and multi-criterion selection plus a mixed strategy of three repair methods) is designed for the multi-objective optimization problem.  Experiment results demonstrate that the MOGA using  helper objectives outperforms its rivals, which are the  greedy algorithm, MSGA and greedy algorithm + MSGA.  The results also show that the  MSGA can find reasonably good solutions but without a guarantee; and the greedy algorithm +
MSGA    sometimes gets trapped into a local optimum.


\end{document}